\documentclass[runningheads]{llncs}

\usepackage{eccv}

\usepackage{eccvabbrv}

\usepackage{graphicx}
\usepackage{booktabs}

\usepackage[accsupp]{axessibility}  

\usepackage{multirow}
\usepackage{comment}
\usepackage{xcolor}

\definecolor{NavyBlue}{HTML}{4169E1}
\definecolor{DarkRed}{HTML}{990000}

\newcommand{\localAug}[1]{{\color{NavyBlue}#1}}
\newcommand{\globalAug}[1]{{\color{DarkRed}#1}}

\newcommand*\samethanks[1][\value{footnote}]{\footnotemark[#1]}

\usepackage[
breaklinks,colorlinks,citecolor=eccvblue]{hyperref}

\usepackage{orcidlink}

\begin{document}

\title{
BackFlip: The Impact of Local and Global Data Augmentations on Artistic Image Aesthetic Assessment
}

\titlerunning{BackFlip: The Impact of Local and Global Data Augmentations
}


\author{Ombretta Strafforello\thanks{Equal contribution.}
\orcidlink{0000-0002-5258-8534} 
\and
Gonzalo Muradas Odriozola\samethanks
\orcidlink{0000-0003-2251-2915} 
\and
Fatemeh Behrad\samethanks
\orcidlink{0000-0003-2629-0854} 
\and
Li-Wei Chen\samethanks
\orcidlink{0009-0003-3116-5836} 
\and
Anne-Sofie Maerten\samethanks
\orcidlink{0000-0003-2710-0936} 
\and
Derya Soydaner\samethanks
\orcidlink{0000-0002-3212-6711} 
\and 
Johan Wagemans
\orcidlink{0000-0002-7970-1541} 
}

\authorrunning{O.~Strafforello et al.}


\institute{
KU Leuven, Leuven, Belgium \\
\url{https://gestaltrevision.be} \\
}

\maketitle

\begin{abstract}
Assessing the aesthetic quality of artistic images presents unique challenges due to the subjective nature of aesthetics and 
the complex visual characteristics inherent to artworks. Basic data augmentation techniques commonly applied to natural images in computer vision may not be suitable for art images in aesthetic evaluation tasks, as they can change the composition of the art images. In this paper, we explore the impact of local and global data augmentation techniques on artistic image aesthetic assessment (IAA). We introduce \emph{BackFlip}, a local data augmentation technique designed specifically for artistic IAA. 
We evaluate the performance of BackFlip across three artistic image datasets and four neural network architectures, comparing it with the commonly used data augmentation techniques. 
Then, we analyze the effects of components within the BackFlip pipeline through an ablation study. Our findings demonstrate that local augmentations, such as BackFlip, tend to outperform global augmentations on artistic IAA in most cases, probably because they do not perturb the composition of the art images. 
These results emphasize the importance of 
considering both local and global augmentations in future computational aesthetics research.

\end{abstract}

\section{Introduction}
\label{sec:intro}

Evaluating image aesthetics is a subjective task for humans, making it even more challenging for neural networks to perform accurately. This task, known as Image Aesthetic Assessment (IAA) in computer science, is part of the interdisciplinary field of computational aesthetics and involves 
modelling aesthetic scores.
The typical approach involves either binary classification, which classifies an image as low or high in aesthetics \cite{talebi2018, lu2014rapid,huang2024coarse,chen2024image}, or regression, which predicts a continuous aesthetic score for a given image \cite{kong2016, ke2021musiq, wang2023exploring, li2023, soydaner2024multi}. In the literature on automated IAA \cite{deng2017,zhai2020}, deep learning plays a crucial role based on its significant impact across various fields. 
However, datasets collected for this task are often limited and struggle to reflect the true aesthetic nature of images. Numerous psychological factors influence aesthetic judgments, resulting in diverse individual preferences. These factors range widely, from low-level image properties and mid-level organizational qualities to high-level semantic content, including individual, social, and cultural factors. Consequently, IAA presents a difficult challenge for artificial intelligence, as it seeks to emulate human aesthetic evaluation. This difficulty is amplified when considering the aesthetic assessment of artworks. Artworks are inherently complex and diverse, characterized by variable compositions and styles (from highly realistic to purely abstract). 
This specific task, known as Artistic IAA, has yet to be fully explored.

Neural network approaches in computer vision commonly use data augmentation techniques to improve performance. However, in artistic IAA, these techniques exhibit limited effectiveness given the importance of overall composition.
When images are modified with data augmentation techniques like cropping, flipping, or color adjustment, the visual aspects that contribute to their aesthetic appeal can be altered. These modifications likely disrupt the original composition, harmony, or emotional impact intended by the artist, invalidating the use of 
the aesthetic scores originally assigned by human participants. 

\begin{figure}[t]
    \centering
    \includegraphics[width=0.9\textwidth]{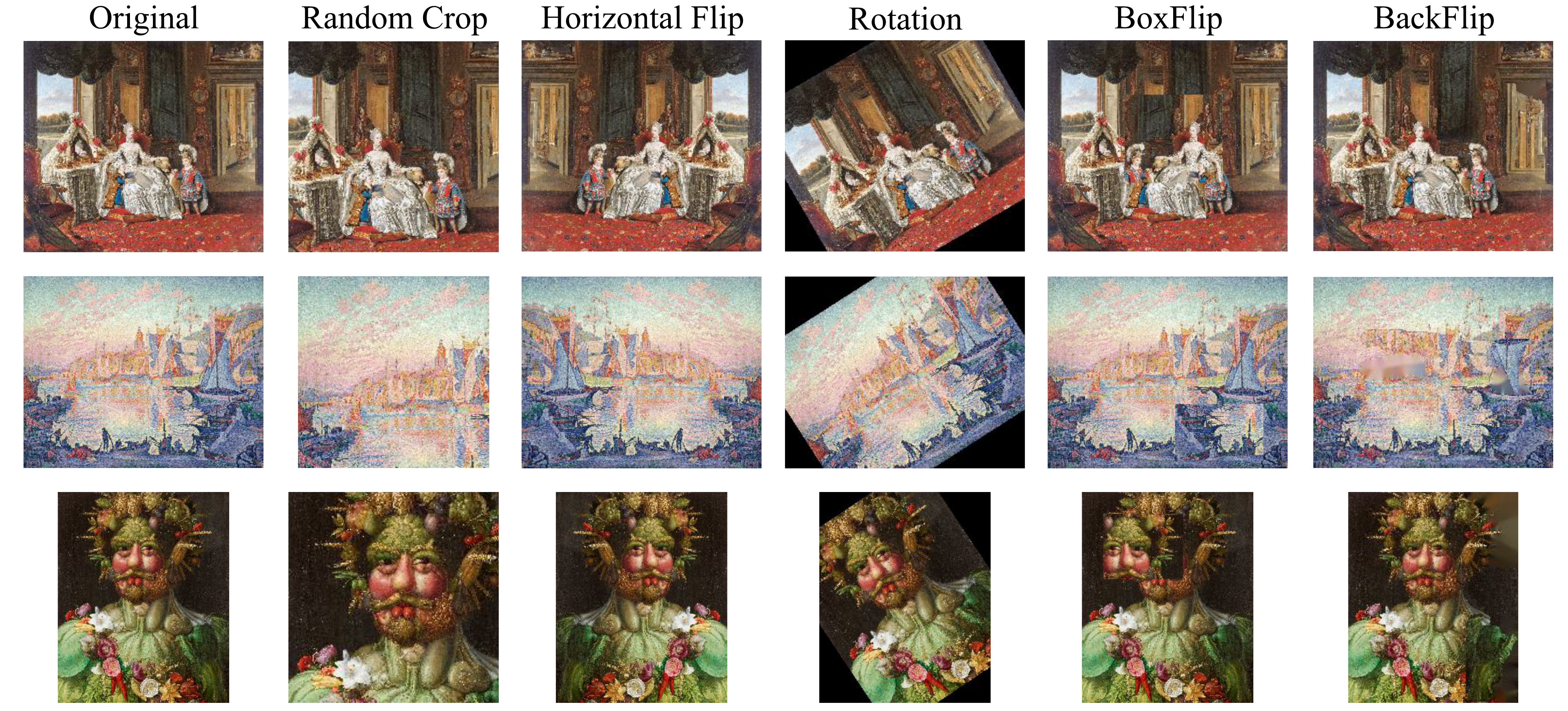}
    \caption{Global (\globalAug{\textit{Random Crop}}, \globalAug{\textit{Horizontal Flip}}, \globalAug{\textit{Rotation}}) and local (\localAug{\textit{BoxFlip}}, \localAug{\textit{BackFlip}}) image transformations on art images from the JenAesthetics dataset \cite{amirshahi2013, amirshahi2013a, amirshahi2014}. The local data augmentations generally preserve
    the global composition of the images, while introducing considerable pixel-level changes that are often less perceptible to the human eye unless if they distort perceptually important shapes and objects like faces.}
    \label{fig:Figure1}
\end{figure}

To address this issue, we examine the effects of data augmentation on artistic IAA. We compare well-known techniques and propose a novel technique called \emph{BackFlip}\footnote{The code is available at \url{https://github.com/GMuradas99/BackFlip}.}, which involves the \emph{local flipping of image regions}. We first segment the images using the Segment Anything (SAM) model \cite{kirillov2023}, then inpaint the \emph{back}ground, and \emph{flip} the selected segment to implement local data augmentation (see Section~\ref{backflipping}). Our approach aims to minimize alterations to the overall composition, thereby preserving human aesthetic appreciation, while effectively modifying the visual patterns crucial for computer vision recognition. In Fig.~\ref{fig:Figure1}, we exemplify global and local image transformations applied to art images, highlighting the distinct impacts of different augmentations. Our study examines the impact of local and global data augmentation techniques on artistic IAA using three benchmark datasets composed of paintings.  We emphasize the challenges of artwork datasets in the context of data augmentation in computer vision. 

\section{Related Work}

\subsection{Artistic Image Aesthetic Assessment}\label{related_work_aiaa}
Early approaches to artistic IAA involve studies that extract features from paintings for classification. For example, Amirshahi \emph{et al.} \cite{amirshahi2017} uses a set of color features in the field of computer vision and image processing, while Li \emph{et al.} \cite{li2009} employs features representing both global and local characteristics of a painting. Additionally, Guo \emph{et al.} \cite{guo2013} evaluates visual complexity of paintings using features that capture both global and local aspects. 

Recent studies include deep learning approaches such as using convolutional neural networks (CNNs) to predict the aesthetics of Chinese ink paintings \cite{zhang2020}. Wilber \emph{et al.} \cite{wilber2017} presented a large-scale dataset of contemporary artworks and used it for artistic style prediction, improving the generality of existing object classifiers, and studying visual domain adaptation. 
More specifically for the artistic IAA task, the Theme-Style-Color Guided Artistic Image Aesthetics Assessment Network (TSC-Net) \cite{wang2023} assesses art images by fusing aesthetic information with image theme, style, and color. Shi \emph{et al.} \cite{SHI2024127434} presented semantic and style based multiple reference learning for artistic and general IAA. Another recent model, the Style-specific Art Assessment
Network (SAAN) \cite{yi2023towards}, evaluates artistic images by combining style-specific and generic aesthetic features. In our study, we adopt deep learning models, including SAAN, to predict aesthetic scores.

\subsection{Image Data Augmentations}\label{image_data_augmentations}
Whether the task involves natural images or artworks, data augmentation techniques are usually necessary for computer vision \cite{kumar2023image}. Limited labeled data can lead to overfitting. 
Additionally, labeling data is time-consuming and expensive. To address overfitting, various generalization techniques have been proposed, such as dropout \cite{srivastava2014dropout} 
and batch normalization \cite{ioffe2015batch}. Among these, data augmentation is the easiest and one of the most common methods to reduce overfitting \cite{krizhevsky2017imagenet}. Basic data augmentation techniques involve image transformations such as rotation, flipping, and cropping. 
In the context of artistic IAA, several studies examine data augmentation techniques. For instance, a stacking
ensemble method for art style recognition has been presented and the effects of data augmentation such as brightness change and rotation have been examined \cite{mastromichalakis2024}. In a similar line of research \cite{yi2023towards}, image augmentations to train self-supervised models have been explored. Both methods use global image transformations.

Different from previous work on data augmentation on art images, we explore a \textit{local} data augmentation strategy that does not alter the overall composition of an artistic image. Local data augmentation has been applied on natural images in various domains but often with highly noticeable visual effects\cite{kim2021local, zhong2020random}. In random erasing \cite{zhong2020random}, for instance, a rectangular part of an image is selected and the pixels are replaced with ImageNet mean values to introduce occlusions. This technique has been shown to complement existing global data augmentation techniques for image classification, object detection, and person re-identification. Another local augmentation technique divides images into rectangular patches and shuffles and augments a selection of patches \cite{kim2021local}. 
This technique has been proposed to exploit the local bias property of CNNs, stating that local augmentations create more diversity relevant to models that extract local features. This approach demonstrates competitive performance with other data augmentation techniques on image classification. Close to this approach, we also suggest \emph{BoxFlip}, where a rectangular patch in an image is augmented. We compare this baseline to our newly proposed BackFlip, which we test under various conditions. We hypothesize that IAA could benefit from local data augmentation more than global augmentation given the importance of composition in artworks for IAA. 
Our novel technique BackFlip extends previous work by locally transforming image \emph{segments} in order to maintain as much of the overall composition as possible.

\section{BackFlip}\label{backflipping} 

\begin{figure}[t]
    \centering
    \includegraphics[width=0.95\textwidth]{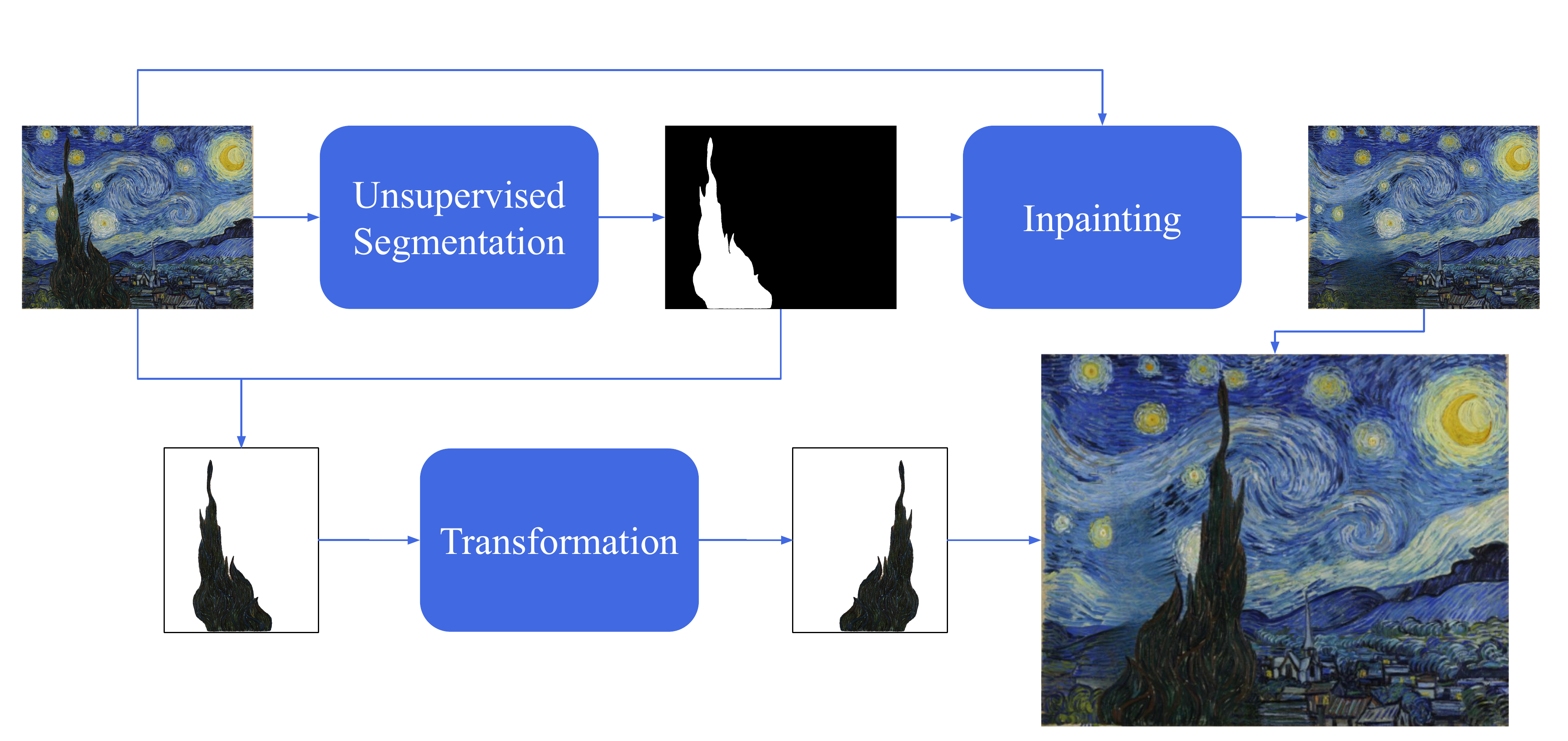}
    \caption{Visualization of the BackFlip pipeline. First, we segment regions in images and inpaint the
    \emph{back}ground, and then we locally \emph{flip} the selected segment to implement data augmentation.
    }
    \label{backPipeline}
\end{figure}

The BackFlip algorithm consists of three primary operations: unsupervised segmentation, inpainting, and local transformations, as shown in 
Figure \ref{backPipeline}. 

\subsection{Segmentation}\label{segmentation_section} 
First, we segment the images using the SAM model \cite{kirillov2023}.  
We use unsupervised segmentation to accommodate the lack of conventional object classes in abstract art.
The SAM implementation in BackFlip returns the segments as binary masks. We exclude segments whose bounding box is larger than 90\% of the image area, given that these segments would alter the overall composition when augmented and can no longer be considered local augmentations (e.g., the background). After excluding those segments, the remaining segments are ordered in descending size. One of the hyperparameters of BackFlip is the number of segments \textit{n} to save. In our tests, SAM detects around 60 segments per artwork on average for all tested datasets. 

It should be noted that classical image augmentation techniques are typically applied during training to introduce various random changes to the data at each epoch. However, BackFlip employs the SAM model for segmentation, which would drastically increase the training time when implemented on each epoch (while yielding the same results every epoch). Therefore, the dataset is pre-segmented once before training to optimize computational resources. During training, segments are selected and locally augmented with a given probability on each epoch, ensuring the model sees a wide variety of augmented images. 

\subsection{Inpainting} In the next step, we erase the chosen segments in the images and inpaint the background. BackFlip employs three types of inpainting methods, with various computational costs and different levels of complexity. These methods typically involve a trade-off between image quality and computational efficiency. We present them in descending order of complexity, starting with methods that produce the most realistic inpainted images.

The first method is LaMa \cite{suvorov2022resolution}, a deep learning model that uses fast Fourier convolutions \cite{chi2020fast}, providing a receptive field that covers the entire image while remaining computationally efficient. This approach is significantly more lightweight compared to other state-of-the-art inpainting models based on generative methods like CoModGAN \cite{zhao2021large} or Stable-Diffusion \cite{Rombach_2022_CVPR}. The model receives the original image and the segmented area as input and returns an image where the segments are erased and inpainted. Since this method repeats the local statistics of the edge around the erased segment, we first dilate the segment mask to ensure the background no longer contains pixels of the removed segment. Similar to SAM, employing LaMa on each epoch would drastically increase the training time while still yielding the same result on each epoch. Therefore, BackFlip pre-inpaints the images based on their pre-computed segmentation masks before training when LaMa is used. The other inpainting methods demand less computational resources and are therefore implemented during training in BackFlip. 

The second group of inpainting methods employs classical computer vision algorithms for efficient real-time data augmentation. We consider two techniques: Fast Marching Method (Telea) \cite{telea2004image}, and fluid dynamics (NS) \cite{bertalmio2001navier}. 
These methods inpaint image pixels using image gradients or the Laplacian. Treating image intensity as an incompressible flow in fluid dynamics, NS transports the image Laplacian as vorticity into the inpainting area. On the other hand, Telea propagates smoothness along image gradients, iteratively inpainting the image by averaging values from neighboring pixels.

Our final inpainting method is based on the mean or median color of the segment boundary. We first compute a dilated version of the segment mask, from which we subtract the original mask. As such, we obtain the segment boundary (or contour), which is then used to calculate its mean or median color to fill the area of the original segment.

\subsection{Local Transformations} In the final step, we introduce local transformations in the images. 
BackFlip offers common data augmentation techniques locally such as vertical and horizontal flipping, random rotation, brightness jitter, downscaling and upscaling. 
The transformed element is then inserted into the inpainted image. 
Augmentations are applied on every epoch with a given probability, which is another hyperparameter that can be adjusted.

\section{Results}

In this section, we evaluate the impact of local and global data augmentations on artistic IAA. We assess the performance of BackFlip across three artistic image datasets and four neural network architectures, comparing it with commonly used data augmentation techniques. The models tested are ResNet-18 \cite{he2016deep}, ResNet-50, ResNeXt-50 \cite{xie2017aggregated}, and SAAN \cite{yi2023towards}. 
We consider a fixed hyperparameter and training setup for each dataset and model combination, with all models pre-trained on ImageNet \cite{deng2009imagenet}. 
We detail the experimental setup in Section~\ref{setup} and present the results in Section~\ref{models}. To ensure robustness and fairness in comparing augmentations, we perform five independent runs for each experiment. Finally, we evaluate the components of the BackFlip pipeline through an ablation study in Section~\ref{ablation}. Our evaluations are based on the Pearson correlation coefficient (PCC) and Spearman's rank correlation coefficient (SRCC) between the ground-truth aesthetic scores of images and the model's predictions. We also assess the classification performance of the models by defining a threshold of 0.5. 

\subsection{Datasets and Experimental Setup}\label{setup}


\subsubsection{BAID.}
The Boldbrush Artistic Image Dataset (BAID) \cite{yi2023towards} consists of 60,337 artistic images covering various art forms, with more than 360,000 votes from online users. Yi \emph{et al.} \cite{yi2023towards} constructed this dataset entirely from artworks obtained from the website Boldbrush\footnote{https://faso.com/boldbrush/popular}. This website hosts a monthly artwork contest where certified artists upload their works and receive public votes from online users. The scores of the images in BAID range from 0 to 10, where 0 represents a lower number of votes and 10 is a higher number of votes. BAID is the most recent IAA dataset and is considered the largest of its kind currently available.

We trained all models for 50 epochs, using various batch sizes depending on the model size (ranging from 50 to 512). All models were trained with a learning rate of 0.001 and the Adam optimizer \cite{kingma2014adam}, except SAAN, trained with 0.0001 and AdamW optimizer \cite{loshchilov2019decoupled}. For the local data augmentations, we selected three segments and used median inpainting, as this method is less costly for the large BAID dataset. We considered horizontal and vertical flips as local augmentations, each with a probability of 0.5. The number of segments \textit{n} to save in BackFlip is 5 for all experiments. 
Pre-segmenting the BAID dataset takes 29 hours, 35 minutes, and 12 seconds for 60337 images on 1 A100 GPU. 

\subsubsection{JenAesthetics.}
The JenAesthetics Subjective Dataset of Aesthetic Paintings \cite{amirshahi2013, amirshahi2013a, amirshahi2014} consists of 1,628 art images. These images are colored oil paintings, all displayed in museums and scanned at high resolution. The dataset covers 11 art periods/styles, including Renaissance, Baroque, and Impressionism, created by 410 artists. 
This dataset provides aesthetic quality scores (how aesthetic the image is) and beauty scores (how beautiful the image is). The rating scale is continuous, ranging from 1 to 100. Additionally, it includes scores for liking of color, content, composition, knowledge of the artist, and familiarity with the painting. Each painting was evaluated by 19-21 observers. Due to some broken URLs in the original dataset, we obtained 1,576 out of the 1,628 images. The train, validation, and test sets consist of 1,103, 158, and 315 images, respectively.

We trained all models for 60 epochs. We trained SAAN with a batch size of 32, using AdamW as the optimizer and a learning rate of 0.0001. The other models were trained with a batch size of 128, using Adam as the optimizer and a learning rate of 0.001. For the local data augmentations, we used Telea as inpainting method. When using BackFlip, we applied the same configuration as used in BAID. 
Pre-segmenting the JenAesthetics dataset takes 1 hour, 49 minutes, and 17 seconds for 1584 images on 1 A4500 Laptop GPU.  

\subsubsection{TAD66K.}
The Theme and Aesthetics Dataset with 66K images (TAD66K) \cite{he2022} is specifically designed for IAA, containing 66k images. 
It covers 47 popular themes, which are the most uploaded on the Flickr website from 2008 to 2021. These themes are grouped into seven superthemes, namely, plants, animals, artifacts, colors, humans, landscapes, and others, which were further divided into 47 subthemes. 
Images of each theme are annotated independently, and each image contains at least 1200 annotations. The annotation score of each image ranges from 1 to 10, representing the lowest aesthetics to the highest aesthetics. They calculated the average value as the ground-truth of the image. 

In our study, we focus on the artistic images within the TAD66K dataset, similar to \cite{SHI2024127434}. This subset contains 1431 labeled artistic images. We maintain the original dataset's split, allocating 289 images for testing and 1,142 images for training. We randomly split the training set, designating 229 images for validation. We trained all models for 100 epochs with a batch size of 128, a learning rate of 0.0001, and the AdamW optimizer. We used median inpainting for our local data augmentations and implemented BackFlip with the same configuration as used in BAID.
Pre-segmenting the TAD66K dataset takes 1 hour, 30 minutes, and 40 seconds for 1431 images on 1 A4500 Laptop GPU. 

\subsection{Artistic IAA Models}\label{models}

To assess the impact of data augmentation on artistic IAA, we consider both global and local augmentation techniques. For the global data augmentation techniques, we include horizontal flip, vertical flip, and rotation. For most augmentations, we first resize every image maintaining the original aspect ratio and reducing the shortest side to 224. Then, we crop part of the image to obtain an input of 224$\times$224. We consider both center cropping and random cropping. Additionally, we consider random cropping without resizing beforehand, referred to as \emph{random resized crop}. We include resize and center crop as a baseline for image preprocessing and report the results before adding augmentation techniques.

For the local data augmentation techniques, 
we implement BoxFlip by selecting a random patch in the image with a min-ratio of 0.3 and a max-ratio of 0.5, which is then flipped either horizontally or vertically. 
To assess the impact of the local image augmentations using BackFlip, we also propose the \emph{erase and inpaint} method, which removes segments and inpaints the background without augmenting the segment.  

Table~\ref{tab:BAID_results} presents the results on the BAID dataset, showing an overall tendency for local data augmentations to perform slightly better than global ones. In terms of accuracy, erase+inpaint performs the best for ResNet-18, followed by BackFlip. In terms of correlations, all local augmentations (erase+inpaint, BoxFlip, and BackFlip) outperform the others in PCC, whereas there is no significant difference in SRCC, except for random resized crop, which comes to the forefront. For ResNet-50, BoxFlip outperforms the others. In ResNeXt-50, random resized crop is the best in terms of accuracy, but local data augmentations provide competitive correlations. We observe a similar trend in SAAN. Additionally, we compare the average results of all global data augmentation experiments with those of all local data augmentation experiments, which is shown in the right 3 columns. This comparison shows that local augmentations perform better than global ones, except in ResNeXt-50, where they perform similarly. 

\begin{table}[ht!]
\fontsize{6.8}{9}\selectfont

\begin{tabular}{ll|lll|lll}
& \multirow{2}{*}{\textbf{Augmentation}} & \multicolumn{6}{c}{\textbf{BAID}} \\
\multicolumn{1}{c}{} &  & \textbf{Acc. (\%)} & \textbf{PCC} & \textbf{SRCC} & \textbf{Acc. (\%)} & \textbf{PCC} & \textbf{SRCC} \\
\midrule 

\multirow{9}{*}{\rotatebox[origin=c]{90}{\textbf{ResNet-18}}} & R., C.C. & $ 74.9 \pm 1.48 $ & $ 0.41 \pm 0.04 $ & $ 0.29 \pm 0.03 $  & & & \\
%
 & \globalAug{R., Random Crop} & $ 70.24 \pm 3.81 $ & $ 0.30 \pm 0.03 $ & $ 0.22 \pm 0.02 $ & \multirow{5}{*}{\globalAug{$ 70.8 \pm 0.2 $}} & \multirow{5}{*}{\globalAug{$ 0.30 \pm 0.0 $}} & \multirow{5}{*}{\globalAug{$ 0.23 \pm 0.01 $}} \\
 & \globalAug{Random Resized Crop} & $ 72.5 \pm 4.64 $ & $ 0.29 \pm 0.06 $ & $ 0.28 \pm 0.05 $  & & & \\
 & \globalAug{R., C.C., Horizontal flip} &  $ 70.31 \pm 1.6 $ & $ 0.31 \pm 0.03 $ & $ 0.21 \pm 0.02 $ & & & \\
 & \globalAug{R., C.C., Vertical flip} &  $ 70.04 \pm 4.09 $ & $ 0.31 \pm 0.02 $ & $ 0.22 \pm 0.02 $ & & & \\
 & \globalAug{R., Rotation, C.C.} &  $ 70.93 \pm 1.94 $ & $ 0.31 \pm 0.03 $ & $ 0.23 \pm 0.03 $ & & & \\
 
 & \localAug{R., C.C., Erase+Inpaint} & $ 73.35 \pm 3.09 $ & $ 0.36 \pm 0.05 $ & $ 0.23 \pm 0.04 $  & \multirow{3}{*}{\localAug{$ 72.47 \pm 0.71 $}} & \multirow{3}{*}{\localAug{$ 0.36 \pm 0.01 $}} & \multirow{3}{*}{\localAug{$ 0.24 \pm 0.01 $}} \\
 & \localAug{R., C.C., BoxFlip} & $ 71.01 \pm 3.19 $ & $ 0.37 \pm 0.03 $ & $ 0.22 \pm 0.06 $ & & & \\
 & \localAug{R., C.C., BackFlip} & $ 72.93 \pm 1.95 $ & $ 0.34 \pm 0.03 $ & $ 0.22 \pm 0.02 $  & & & \\
\midrule

\multirow{9}{*}{\rotatebox[origin=c]{90}{\textbf{ResNet-50}}} & R., C.C. & $ 70.14 \pm 1.88 $ & $ 0.29 \pm 0.05 $ & $ 0.2 \pm 0.05 $ & & & \\
%
 & \globalAug{R., Random Crop} & $ 72.29 \pm 1.26 $ & $ 0.29 \pm 0.04 $ & $ 0.22 \pm 0.03 $ & \multirow{5}{*}{\globalAug{$ 71.97 \pm 0.11 $}} & \multirow{5}{*}{\globalAug{$ 0.30 \pm 0.0 $}} & \multirow{5}{*}{\globalAug{$ 0.24 \pm 0.0 $}} \\
 & \globalAug{Random Resized Crop} & $ 72.58 \pm 5.64 $ & $ 0.28 \pm 0.08 $ & $ 0.25 \pm 0.05 $ & & & \\
 & \globalAug{R., C.C., Horizontal flip} & $ 71.15 \pm 5.7 $ & $ 0.32 \pm 0.05 $ & $ 0.24 \pm 0.05 $ & & & \\
 & \globalAug{R., C.C., Vertical flip} &  $ 72.17 \pm 0.85 $ & $ 0.33 \pm 0.05 $ & $ 0.25 \pm 0.05 $ & & & \\
 & \globalAug{R., Rotation, C.C.} & $ 71.69 \pm 4.84 $ & $ 0.3 \pm 0.01 $ & $ 0.24 \pm 0.02 $ & & & \\
 
 & \localAug{R., C.C., Erase+Inpaint} & $ 72.59 \pm 2.72 $ & $ 0.36 \pm 0.03 $ & $ 0.23 \pm 0.04 $  &  \multirow{3}{*}{\localAug{$ 73.37 \pm 0.28 $}} & \multirow{3}{*}{\localAug{$ 0.37 \pm 0.0 $}} & \multirow{3}{*}{\localAug{$ 0.25 \pm 0.0 $}} \\
 & \localAug{R., C.C., BoxFlip} & $ 74.02 \pm 0.6 $ & $ 0.38 \pm 0.02 $ & $ 0.24 \pm 0.02 $ & & & \\
 & \localAug{R., C.C., BackFlip} & $ 70.48 \pm 4.57 $ & $ 0.33 \pm 0.05 $ & $ 0.22 \pm 0.04 $  & & & \\
\midrule

\multirow{8}{*}{\rotatebox[origin=c]{90}{\textbf{ResNeXt50}}} & R., C.C. &  $ 73.3 \pm 3.13 $ & $ 0.37 \pm 0.05 $ & $ 0.27 \pm 0.02 $ &  & &   \\
%
 & \globalAug{R., Random Crop} & $ 74.49 \pm 1.03 $ & $ 0.32 \pm 0.09 $ & $ 0.25 \pm 0.09 $ & \multirow{5}{*}{\globalAug{$ 73.6 \pm 0.24 $}} & \multirow{5}{*}{\globalAug{$ 0.30 \pm 0.01 $}} & \multirow{5}{*}{\globalAug{$ 0.25 \pm 0.01 $}} \\
 & \globalAug{Random Resized Crop} & $ 75.13 \pm 2.33 $ & $ 0.3 \pm 0.13 $ & $ 0.32 \pm 0.04 $ & & & \\
 & \globalAug{R., C.C., Horizontal flip} & $ 73.48 \pm 0.82 $ & $ 0.34 \pm 0.05 $ & $ 0.25 \pm 0.04 $ & & & \\
 & \globalAug{R., C.C., Vertical flip} &  $ 72.6 \pm 1.65 $ & $ 0.3 \pm 0.02 $ & $ 0.23 \pm 0.01 $ & & & \\
 & \globalAug{R., Rotation, C.C.} & $ 72.32 \pm 2.26 $ & $ 0.24 \pm 0.14 $ & $ 0.2 \pm 0.07 $   &  &  &  \\
 
 & \localAug{R., C.C., Erase+Inpaint} &   $ 73.86 \pm 1.25 $ & $ 0.35 \pm 0.03 $ & $ 0.25 \pm 0.01 $   &  \multirow{3}{*}{\localAug{$ 73.51 \pm 0.22 $}} & \multirow{3}{*}{\localAug{$ 0.34 \pm 0.01 $}} & \multirow{3}{*}{\localAug{$ 0.24 \pm 0.0 $}} \\
 & \localAug{R., C.C., BoxFlip} & $ 73.93 \pm 1.35 $ & $ 0.35 \pm 0.03 $ & $ 0.23 \pm 0.02 $ & & & \\
 & \localAug{R., C.C., BackFlip} & $ 72.74 \pm 2.02 $ & $ 0.32 \pm 0.03 $ & $ 0.23 \pm 0.02 $ & & & \\
\midrule

\multirow{8}{*}{\rotatebox[origin=c]{90}{\textbf{SAAN}}} & R., C.C. & $ 74.77 \pm 1.37 $ & $ 0.37 \pm 0.03 $ & $ 0.3 \pm 0.04 $ &  & & \\
%
 & \globalAug{R., Random Crop} & $ 72.4 \pm 1.16 $ & $ 0.37 \pm 0.03 $ & $ 0.29 \pm 0.03 $ & \multirow{5}{*}{\globalAug{$ 73.43 \pm 0.37 $}} & \multirow{5}{*}{\globalAug{$ 0.38 \pm 0.0 $}} & \multirow{5}{*}{\globalAug{$ 0.32 \pm 0.01 $}} \\
 & \globalAug{Random Resized Crop} & $ 76.46 \pm 0.37 $ & $ 0.41 \pm 0.03 $ & $ 0.37 \pm 0.02 $ & & & \\
 & \globalAug{R., C.C., Horizontal flip} & $ 72.4 \pm 1.7 $ & $ 0.37 \pm 0.04 $ & $ 0.31 \pm 0.04 $   & & & \\
 
 & \globalAug{R., C.C., Vertical flip} & $ 73.95 \pm 0.8 $ & $ 0.35 \pm 0.02 $ & $ 0.3 \pm 0.03 $ & & & \\
 & \globalAug{R., Rotation, C.C.} & $ 71.95 \pm 1.59 $ & $ 0.39 \pm 0.01 $ & $ 0.31 \pm 0.01 $   &  &  &  \\
 
 & \localAug{R., C.C., Erase+Inpaint} &  $ 74.09 \pm 1.01 $ & $ 0.41 \pm 0.02 $ & $ 0.32 \pm 0.03 $  & \multirow{3}{*}{\localAug{$ 73.61 \pm 0.14 $}} & \multirow{3}{*}{\localAug{$ 0.39 \pm 0.01 $}} & \multirow{3}{*}{\localAug{$ 0.31 \pm 0.01 $}} \\
 & \localAug{R., C.C., BoxFlip} &  $ 73.27 \pm 1.74 $ & $ 0.37 \pm 0.04 $ & $ 0.29 \pm 0.03 $ & & & \\
 & \localAug{R., C.C., BackFlip} & $ 73.47 \pm 1.05 $ & $ 0.4 \pm 0.01 $ & $ 0.33 \pm 0.02 $   & & & \\
\midrule 

\end{tabular}
\caption{Results on BAID for four 
models trained with different \globalAug{global} and \localAug{local} data augmentation techniques. \textit{R.} and \textit{C.C.} 
stand, respectively, for \textit{Resize} and \textit{Center Crop}. 
We report the average accuracy across at least 5 independent runs. The ensemble mean and standard deviation for \globalAug{global} and \localAug{local} data augmentations are displayed in the right column. For the local data augmentations, we used median inpainting.
%
}
\label{tab:BAID_results}
\end{table}

We repeat these experiments on the other datasets in our study, with Table~\ref{tab:JenAesthetics_results} showing the results for the JenAesthetics dataset. We observe a similar tendency as in the previous results, but it is more evident. 
In terms of SRCC, the average performance of all local data augmentations outperforms that of global data augmentations in ResNet-18 and ResNet-50. When comparing accuracy across all models, local augmentations usually outperform or perform similarly to global ones. In terms of correlations, BackFlip is superior to the other techniques in SAAN.
We also emphasize that JenAesthetics is a better-curated dataset of paintings compared to the others. 

\begin{table}[!ht]
\fontsize{6.8}{9}\selectfont

\begin{tabular}{ll|lll|lll}
& \multirow{2}{*}{\textbf{Augmentation}} & \multicolumn{6}{c}{\textbf{JenAesthetics}} \\
\multicolumn{1}{c}{} &  & \textbf{Acc. (\%)} & \textbf{PCC} & \textbf{SRCC} & \textbf{Acc. (\%)} & \textbf{PCC} & \textbf{SRCC} \\
\midrule

\multirow{9}{*}{\rotatebox[origin=c]{90}{\textbf{ResNet-18}}} 
& R., C.C.                   &  $ 71.43 \pm 2.31 $ & $ 0.25 \pm 0.03 $ & $ 0.24 \pm 0.02 $  & \\
%
& \globalAug{R., Random Crop} & $ 71.3 \pm 1.58 $ & $ 0.16 \pm 0.02 $ & $ 0.14 \pm 0.02 $ & \multirow{5}{*}{\globalAug{$ 70.9 \pm 0.6 $}} & \multirow{5}{*}{\globalAug{$ 0.17 \pm 0.01 $}} & \multirow{5}{*}{\globalAug{$ 0.15 \pm 0.01 $}} \\
& \globalAug{Random Resized Crop} & $ 73.49 \pm 2.5 $ & $ 0.23 \pm 0.03 $ & $ 0.2 \pm 0.04 $  &  & & \\
& \globalAug{R., C.C., Horizontal flip}  & $ 73.78 \pm 1.48 $ & $ 0.13 \pm 0.02 $ & $ 0.09 \pm 0.02 $  & & & \\
& \globalAug{R., C.C., Vertical flip} & $ 69.27 \pm 2.06 $ & $ 0.15 \pm 0.0 $ & $ 0.16 \pm 0.01 $  & & &\\
& \globalAug{R., Rotation, C.C.} & $ 66.67 \pm 1.63 $ & $ 0.16 \pm 0.04 $ & $ 0.15 \pm 0.03 $  & & & \\


& \localAug{R., C.C., Erase+Inpaint}  & $ 71.81 \pm 1.7 $ & $ 0.2 \pm 0.02 $ & $ 0.18 \pm 0.02 $  &  \multirow{3}{*}{\localAug{$ 69.00 \pm 2.54 $}} & \multirow{3}{*}{\localAug{$ 0.19 \pm 0.02 $}} & \multirow{3}{*}{\localAug{$ 0.19 \pm 0.02 $}} \\ 

& \localAug{R., C.C., BoxFlip}         & $ 69.53 \pm 1.85 $ & $ 0.22 \pm 0.01 $ & $ 0.21 \pm 0.01 $  & & & \\
& \localAug{R., C.C., BackFlip}        & $ 65.65 \pm 2.22 $ & $ 0.16 \pm 0.02 $ & $ 0.17 \pm 0.01 $  & & & \\ 
\midrule

\multirow{9}{*}{\rotatebox[origin=c]{90}{\textbf{ResNet-50}}} 
& R., C.C.                   & $ 75.62 \pm 0.57 $ & $ 0.17 \pm 0.02 $ & $ 0.15 \pm 0.02 $  & & &  \\
%
& \globalAug{R., Random Crop} & $ 75.56 \pm 0.87 $ & $ 0.22 \pm 0.02 $ & $ 0.19 \pm 0.02 $ & \multirow{5}{*}{\globalAug{$ 72.39 \pm 0.61 $}} & \multirow{5}{*}{\globalAug{$ 0.16 \pm 0.01 $}} & \multirow{5}{*}{\globalAug{$ 0.14 \pm 0.01 $}} \\
& \globalAug{Random Resized Crop} & $ 75.87 \pm 0.5 $ & $ 0.17 \pm 0.03 $ & $ 0.15 \pm 0.03 $ & & & \\
& \globalAug{R., C.C., Horizontal flip}  & $ 70.48 \pm 2.51 $ & $ 0.17 \pm 0.04 $ & $ 0.14 \pm 0.04 $   & & & \\
& \globalAug{R., C.C., Vertical flip} & $ 69.71 \pm 2.25 $ & $ 0.13 \pm 0.02 $ & $ 0.13 \pm 0.02 $ &  & & \\
& \globalAug{R., Rotation, C.C.} & $ 70.35 \pm 2.55 $ & $ 0.13 \pm 0.04 $ & $ 0.09 \pm 0.03 $ 
& & &\\

 
& \localAug{R., C.C., Erase+Inpaint} &  $ 75.81 \pm 0.73 $ & $ 0.19 \pm 0.03 $ & $ 0.18 \pm 0.02 $  &  \multirow{3}{*}{\localAug{$ 74.96 \pm 0.60 $}} & \multirow{3}{*}{\localAug{$ 0.20 \pm 0.01 $}} & \multirow{3}{*}{\localAug{$ 0.18 \pm 0.03 $}} \\ 

& \localAug{R., C.C., BoxFlip}         & $ 74.48 \pm 0.73 $ & $ 0.19 \pm 0.02 $ & $ 0.15 \pm 0.02 $   & & & \\
& \localAug{R., C.C., BackFlip}     & $ 74.60 \pm 1.90 $ & $ 0.22 \pm 0.05 $ & $ 0.22 \pm 0.05 $   & & & \\ 
\midrule

\multirow{9}{*}{\rotatebox[origin=c]{90}{\textbf{ResNeXt50}}}
& R., C.C.                   & $ 69.33 \pm 3.16 $ & $ 0.14 \pm 0.07 $ & $ 0.13 \pm 0.03 $  & & & \\
%
& \globalAug{R., Random Crop} & $ 71.55 \pm 4.77 $ & $ 0.19 \pm 0.05 $ & $ 0.18 \pm 0.04 $ & \multirow{5}{*}{\globalAug{$ 72.15 \pm 0.5 $}} & \multirow{5}{*}{\globalAug{$ 0.19 \pm 0.01 $}} & \multirow{5}{*}{\globalAug{$ 0.17 \pm 0.01 $}} \\
& \globalAug{Random Resized Crop} & $ 75.81 \pm 1.41 $ & $ 0.26 \pm 0.03 $ & $ 0.22 \pm 0.04 $ & & & \\
& \globalAug{R., C.C., Horizontal flip}  & $ 68.83 \pm 1.46 $ & $ 0.16 \pm 0.07 $ & $ 0.15 \pm 0.04 $  & & & \\
& \globalAug{R., C.C., Vertical flip} & $ 72.83 \pm 1.7 $ & $ 0.2 \pm 0.03 $ & $ 0.17 \pm 0.02 $ & & &\\
& \globalAug{R., Rotation, C.C.} & $ 71.75 \pm 2.88 $ & $ 0.13 \pm 0.03 $ & $ 0.13 \pm 0.03 $ & & & \\

& \localAug{R., C.C., Erase+Inpaint}    & $ 75.62 \pm 2.08 $ & $ 0.19 \pm 0.02 $ & $ 0.16 \pm 0.02 $  &  \multirow{3}{*}{\localAug{$ 75.13 \pm 0.34 $}} & \multirow{3}{*}{\localAug{$ 0.17 \pm 0.04 $}} & \multirow{3}{*}{\localAug{$ 0.15 \pm 0.03 $}}  \\ %
& \localAug{R., C.C., BoxFlip}          & $ 74.92 \pm 2.74 $ & $ 0.12 \pm 0.03 $ & $ 0.11 \pm 0.04 $ & & & \\
& \localAug{R., C.C., BackFlip}        & $ 74.86 \pm 0.73 $ & $ 0.20 \pm 0.04 $ & $ 0.17 \pm 0.06 $  & & & \\ 
\midrule

\multirow{9}{*}{\rotatebox[origin=c]{90}{\textbf{SAAN}}} 
& R., C.C.                   &$ 73.52 \pm 1.72 $ & $ 0.18 \pm 0.05 $ & $ 0.22 \pm 0.05 $   & & & \\
%
& \globalAug{R., Random Crop} & $ 75.43 \pm 1.04 $ & $ 0.29 \pm 0.02 $ & $ 0.28 \pm 0.02 $  & \multirow{5}{*}{\globalAug{$ 75.12 \pm 0.08 $}} & \multirow{5}{*}{\globalAug{$ 0.25 \pm 0.01 $}} & \multirow{5}{*}{\globalAug{$ 0.25 \pm 0.02 $}} \\
& \globalAug{Random Resized Crop} & $ 75.49 \pm 1.41 $ & $ 0.25 \pm 0.06 $ & $ 0.26 \pm 0.04 $ & & & \\
& \globalAug{R., C.C., Horizontal flip}  & $ 74.54 \pm 1.09 $ & $ 0.2 \pm 0.03 $ & $ 0.22 \pm 0.02 $  & & & \\
& \globalAug{R., C.C., Vertical flip} & $ 75.11 \pm 1.09 $ & $ 0.26 \pm 0.02 $ & $ 0.24 \pm 0.02 $  & & & \\
& \globalAug{R., Rotation, C.C.} & $ 75.05 \pm 1.73 $ & $ 0.27 \pm 0.03 $ & $ 0.25 \pm 0.02 $  & & & \\
 
& \localAug{R., C.C., Erase+Inpaint} & $ 74.6 \pm 1.92 $ & $ 0.24 \pm 0.04 $ & $ 0.22 \pm 0.03 $ &  \multirow{3}{*}{\localAug{$ 74.64 \pm 0.21 $}} & \multirow{3}{*}{\localAug{$ 0.26 \pm 0.04 $}} & \multirow{3}{*}{\localAug{$ 0.25 \pm 0.03 $}}  \\ 
& \localAug{R., C.C., BoxFlip}          & $ 74.92 \pm 2.36 $ & $ 0.22 \pm 0.08 $ & $ 0.23 \pm 0.06 $  & & & \\
& \localAug{R., C.C., BackFlip}  & $ 74.41 \pm 0.86 $ & $ 0.31 \pm 0.02 $ & $ 0.29 \pm 0.02 $  & & & \\ 
\midrule 

\end{tabular}
\caption{Results on JenAesthetics for different models trained with \globalAug{global} and \localAug{local} image data augmentation techniques. \textit{R.} and \textit{C.C.} stand, respectively, for \textit{Resize} and \textit{Center Crop}. 
We report the average accuracy across 5 independent runs. For the local data augmentations, we used Telea inpainting.
The table format follows that of Table~\ref{tab:BAID_results}.
}
\label{tab:JenAesthetics_results}
\end{table}

Lastly, Table~\ref{tab:aug_accuracies} shows the results on the TAD66K dataset, presenting similar results between the average local and data augmentations. 
In terms of accuracy, PCC, and SRCC, we observe that rotation outperforms the others in ResNet-18 and ResNet-50, with local augmentations following closely. In the ResNeXt-50 and SAAN, we observe similar results between the local augmentations and the others. However, it is important to note that TAD66K-Art subset includes a diverse set of images, such as a picture of two paintings and an observer (see Fig.~\ref{fig:tad66k_inp}, the second row).



\begin{table}[!ht]
\fontsize{6.8}{9}\selectfont

\begin{tabular}{ll|lll|lll}
& \multirow{2}{*}{\textbf{Augmentation}} & \multicolumn{6}{c}{\textbf{TAD66K (artwork)}} \\
\multicolumn{1}{c}{} &  & \textbf{Acc. (\%)} & \textbf{PCC} & \textbf{SRCC} & \textbf{Acc. (\%)} & \textbf{PCC} & \textbf{SRCC} \\
\midrule 

\multirow{9}{*}{\rotatebox[origin=c]{90}{\textbf{ResNet-18}}} & R., C.C. & $ 52.94 \pm 1.83 $ & $ 0.1 \pm 0.07 $ & $ 0.1 \pm 0.07 $  & & & \\

 & \globalAug{R., Random Crop} & $ 57.58 \pm 2.14 $ & $ 0.24 \pm 0.05 $ & $ 0.23 \pm 0.04 $ & \multirow{5}{*}{\globalAug{$ 54.88 \pm 0.42 $}} & \multirow{5}{*}{\globalAug{$ 0.22 \pm 0.01 $}} & \multirow{5}{*}{\globalAug{$ 0.20 \pm 0.01 $}} \\
 & \globalAug{Random Resized Crop} & $ 53.77 \pm 3.87 $ & $ 0.25 \pm 0.03 $ & $ 0.22 \pm 0.04 $ & & & \\
 & \globalAug{R., C.C., Horizontal flip} &  $ 53.83 \pm 3.49 $ & $ 0.17 \pm 0.05 $ & $ 0.15 \pm 0.06 $  & & & \\
 & \globalAug{R., C.C., Vertical flip} &  $ 52.6 \pm 3.1 $ & $ 0.18 \pm 0.04 $ & $ 0.16 \pm 0.04 $ & & & \\
 & \globalAug{R., Rotation, C.C.} & $ 56.61 \pm 2.96 $ & $ 0.27 \pm 0.03 $ & $ 0.25 \pm 0.04 $ & & & \\
 
 & \localAug{R., C.C., Erase+Inpaint}  & $ 53.63 \pm 6.03 $ & $ 0.22 \pm 0.04 $ & $ 0.18 \pm 0.05 $  &  \multirow{3}{*}{\localAug{$ 54.21 \pm 0.28 $}} & \multirow{3}{*}{\localAug{$ 0.23 \pm 0.0 $}} & \multirow{3}{*}{\localAug{$ 0.19 \pm 0.0 $}}  \\
 & \localAug{R., C.C., BoxFlip} & $ 53.84 \pm 1.98 $ & $ 0.24 \pm 0.04 $ & $ 0.19 \pm 0.03 $  & & & \\
 & \localAug{R., C.C., BackFlip} & $ 55.16 \pm 2.7 $ & $ 0.22 \pm 0.03 $ & $ 0.2 \pm 0.03 $  & & & \\
\midrule

\multirow{9}{*}{\rotatebox[origin=c]{90}{\textbf{ResNet-50}}} & R., C.C. &  $ 58.83 \pm 2.75 $ & $ 0.3 \pm 0.03 $ & $ 0.27 \pm 0.03 $  & & &   \\
 & \globalAug{R., Random Crop} & $ 61.39 \pm 3.02 $ & $ 0.32 \pm 0.02 $ & $ 0.3 \pm 0.01 $  &  \multirow{5}{*}{\globalAug{$ 61.07 \pm 0.28 $}} & \multirow{5}{*}{\globalAug{$ 0.33 \pm 0.01 $}} & \multirow{5}{*}{\globalAug{$ 0.31 \pm 0.01 $}} \\
 & \globalAug{Random Resized Crop} & $ 60.76 \pm 3.08 $ & $ 0.3 \pm 0.03 $ & $ 0.3 \pm 0.02 $ & & & \\
 & \globalAug{R., C.C., Horizontal flip} &   $ 59.93 \pm 3.57 $ & $ 0.33 \pm 0.04 $ & $ 0.3 \pm 0.05 $  &  \\
 & \globalAug{R., C.C., Vertical flip} & $ 59.93 \pm 1.71 $ & $ 0.3 \pm 0.03 $ & $ 0.28 \pm 0.02 $  & & & \\
 & \globalAug{R., Rotation, C.C.} & $ 63.32 \pm 1.45 $ & $ 0.38 \pm 0.01 $ & $ 0.38 \pm 0.02 $  & & & \\
 
 & \localAug{R., C.C., Erase+Inpaint}  &  $ 61.53 \pm 1.37 $ & $ 0.34 \pm 0.02 $ & $ 0.3 \pm 0.01 $  &  \multirow{3}{*}{\localAug{$ 60.72 \pm 0.47 $}} & \multirow{3}{*}{\localAug{$ 0.34 \pm 0.0 $}} & \multirow{3}{*}{\localAug{$ 0.32 \pm 0.0 $}} \\
 & \localAug{R., C.C., BoxFlip} & $ 61.52 \pm 1.98 $ & $ 0.33 \pm 0.02 $ & $ 0.32 \pm 0.02 $  & & & \\
 & \localAug{R., C.C., BackFlip} & $ 59.1 \pm 1.65 $ & $ 0.35 \pm 0.03 $ & $ 0.32 \pm 0.03 $  & & & \\
\midrule 

\multirow{9}{*}{\rotatebox[origin=c]{90}{\textbf{ResNeXt50}}} & R., C.C. & $ 59.17 \pm 3.69 $ & $ 0.37 \pm 0.04 $ & $ 0.34 \pm 0.05 $  &   & &  \\ 

 & \globalAug{R., Random Crop} & $ 62.77 \pm 4.54 $ & $ 0.38 \pm 0.02 $ & $ 0.37 \pm 0.01 $  &  \multirow{5}{*}{\globalAug{$ 59.62 \pm 0.56 $}} & \multirow{5}{*}{\globalAug{$ 0.38 \pm 0.0 $}} & \multirow{5}{*}{\globalAug{$ 0.35 \pm 0.0 $}} \\
 & \globalAug{Random Resized Crop} & $ 60.49 \pm 2.83 $ & $ 0.37 \pm 0.03 $ & $ 0.35 \pm 0.02 $ & & & \\
 & \globalAug{R., C.C., Horizontal flip} &  $ 55.25 \pm 6.9 $ & $ 0.39 \pm 0.03 $ & $ 0.36 \pm 0.02 $ &   \\
 & \globalAug{R., C.C., Vertical flip} & $ 60.55 \pm 2.93 $ & $ 0.36 \pm 0.03 $ & $ 0.34 \pm 0.02 $   \\
 & \globalAug{R., Rotation, C.C.} & $ 59.03 \pm 2.08 $ & $ 0.38 \pm 0.04 $ & $ 0.35 \pm 0.03 $ & & & \\
 
 & \localAug{R., C.C., Erase+Inpaint}  & $ 59.03 \pm 1.5 $ & $ 0.38 \pm 0.02 $ & $ 0.34 \pm 0.02 $  & \multirow{3}{*}{\localAug{$ 59.61 \pm 0.18 $}} & \multirow{3}{*}{\localAug{$ 0.38 \pm 0.0 $}} & \multirow{3}{*}{\localAug{$ 0.34 \pm 0.0 $}} \\
 & \localAug{R., C.C., BoxFlip} & $ 60.07 \pm 2.97 $ & $ 0.39 \pm 0.05 $ & $ 0.36 \pm 0.05 $  \\
 & \localAug{R., C.C., BackFlip} &  $ 59.72 \pm 1.13 $ & $ 0.37 \pm 0.05 $ & $ 0.33 \pm 0.06 $  \\
\midrule

\multirow{9}{*}{\rotatebox[origin=c]{90}{\textbf{SAAN}}} &R., C.C. & $ 59.72 \pm 2.68 $ & $ 0.26 \pm 0.05 $ & $ 0.28 \pm 0.03 $   & & &  \\
 
 & \globalAug{R., Random Crop} & $ 59.31 \pm 2.79 $ & $ 0.22 \pm 0.06 $ & $ 0.26 \pm 0.03 $  & \multirow{5}{*}{\globalAug{$ 59.29 \pm 0.13 $}} & \multirow{5}{*}{\globalAug{$ 0.21 \pm 0.01 $}} & \multirow{5}{*}{\globalAug{$ 0.26 \pm 0.0 $}} \\
 & \globalAug{Random Resized Crop} & $ 59.58 \pm 2.33 $ & $ 0.21 \pm 0.06 $ & $ 0.27 \pm 0.03 $  & & & \\
 & \globalAug{R., C.C., Horizontal flip} &   $ 58.69 \pm 1.95 $ & $ 0.21 \pm 0.09 $ & $ 0.24 \pm 0.06 $   &  \\
 & \globalAug{R., C.C., Vertical flip} &  $ 58.69 \pm 1.67 $ & $ 0.17 \pm 0.07 $ & $ 0.24 \pm 0.03 $   & & & \\
 & \globalAug{R., Rotation, C.C.} & $ 60.21 \pm 2.54 $ & $ 0.25 \pm 0.06 $ & $ 0.28 \pm 0.04 $ & & & \\
 
 & \localAug{R., C.C., Erase+Inpaint}   & $ 58.13 \pm 1.2 $ & $ 0.26 \pm 0.06 $ & $ 0.26 \pm 0.03 $  &   \multirow{3}{*}{\localAug{$ 59.03 \pm 0.41 $}} & \multirow{3}{*}{\localAug{$ 0.25 \pm 0.01 $}} & \multirow{3}{*}{\localAug{$ 0.27 \pm 0.0 $}} \\
 & \localAug{R., C.C., BoxFlip} & $ 60.41 \pm 2.62 $ & $ 0.23 \pm 0.06 $ & $ 0.27 \pm 0.05 $  & & & \\
 & \localAug{R., C.C., BackFlip} & $ 58.55 \pm 2.21 $ & $ 0.26 \pm 0.03 $ & $ 0.26 \pm 0.03 $  & & & \\
 \midrule 
 
\end{tabular}
\caption{Results on TAD66K - Art for different models trained with \globalAug{global} and \localAug{local} image data augmentation techniques. \textit{R.} and \textit{C.C.} stand, respectively, for \textit{Resize} and \textit{Center Crop}. 
We report the average accuracy across 5 independent runs. For the local data augmentations, we used median inpainting. The table format follows that of Table~\ref{tab:BAID_results}.
}
\label{tab:aug_accuracies}
\end{table}

\subsection{BackFlip Ablation Study}

We observed that BackFlip, as well as the erase+inpaint and BoxFlip techniques, perform well. Here, we evaluate the design choices of BackFlip through an ablation study. Specifically, we test the impact of the inpainting method, the number of locally augmented segments, and the type of local segment augmentations.

\subsubsection{Inpainting method.} 
\begin{figure}[!ht]
    \centering
    \includegraphics[width=0.9\textwidth]{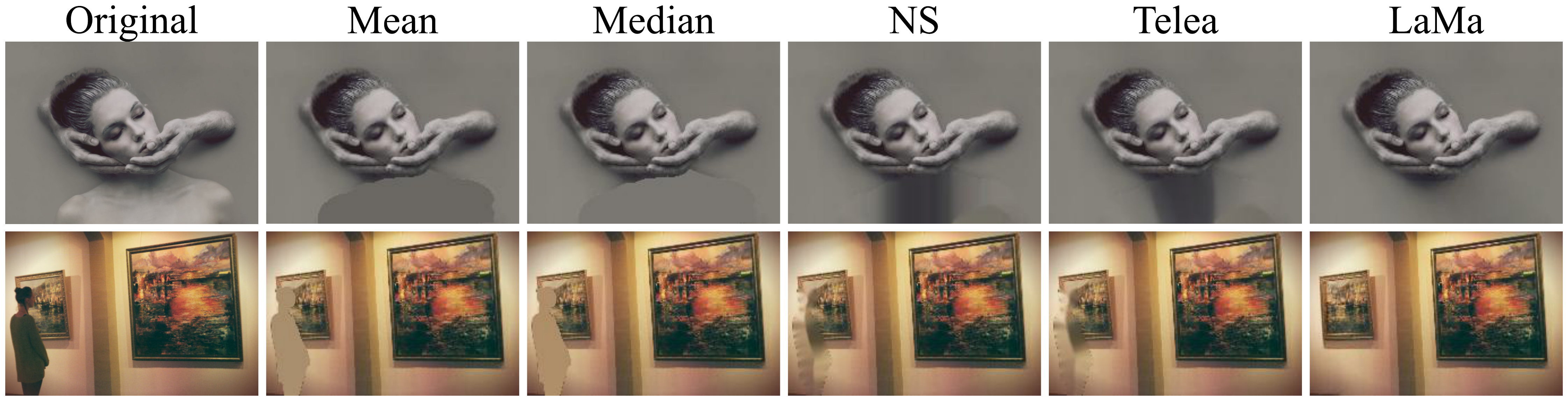}
    \caption{
    Erase + Inpaint (BackFlip without inserting a segmented image) with different inpainting methods on images from TAD66K - Art.}
    \label{fig:tad66k_inp}
\end{figure}

The inpainting component in BackFlip (Fig.~\ref{backPipeline}) can be crucial in the visual quality of the output image while it is unclear whether the models benefit from improved inpainting methods. 
To showcase the inpainting technique in isolation, we tested BackFlip without inserting a segmented image, which is equivalent to 'Erase + Inpaint' method, in Figure~\ref{fig:tad66k_inp}. We compared mean, median, NS, Telea, and LaMa inpainting techniques on two example images from TAD66K - Art. The original images in the first column of this figure also exemplify the diversity of images in the `art' category in TAD66K. One example is a museum picture showing two paintings from an angle, with a visitor partially occluding one painting. The image as a whole is not a painting; segmenting out the person does more than changing the `artwork'.

To compare the inpainting techniques, we train SAAN \cite{yi2023towards} for artistic IAA on the TAD66K dataset. Table~\ref{tab:inpainting} shows the results, keeping the BackFlip setup constant. In this case, we augmented three local segments and used horizontal flipping as the local augmentation. The more refined inpainting techniques, such as LaMa and Telea, yield slightly better results than the others. However, given their higher computational cost and relatively small performance gain, we argue that a simpler approach, such as median inpainting, is a more optimal choice, especially when training on larger datasets.

\begin{table}[!ht]
\fontsize{7}{9}\selectfont
\centering
\begin{tabular}{l|lll}
\hline 
\textbf{Inpainting method} & \textbf{Acc. (\%)} & \textbf{PCC} & \textbf{SRCC} \\
\midrule 
Mean & $56.33 \pm 1.05$ & $0.28 \pm 0.05$ & $0.26 \pm 0.04$ \\
Median & $57.02 \pm 0.79$ & $0.32 \pm 0.02$ & $0.28 \pm 0.02$ \\
Telea & $\mathbf{59.03 \pm 1.46}$ & $0.33 \pm 0.02$ & $0.31 \pm 0.03$ \\
NS & $56.81 \pm 2.22$ & $0.32 \pm 0.03$ & $0.28 \pm 0.03$ \\
LaMa & $58.48 \pm 1.49$ & $\mathbf{0.34 \pm 0.04}$ & $\mathbf{0.32 \pm 0.03}$ \\
\hline 
\end{tabular}
\caption{Testing the effect of BackFlip with different inpainting methods on TAD66K - Art using SAAN \cite{yi2023towards}. We report the average across 5 independent runs.
}
\label{tab:inpainting}
\end{table}

\subsubsection{Local augmentation types.} 
\begin{figure}[!ht]
    \centering
    \includegraphics[width=0.9\textwidth]{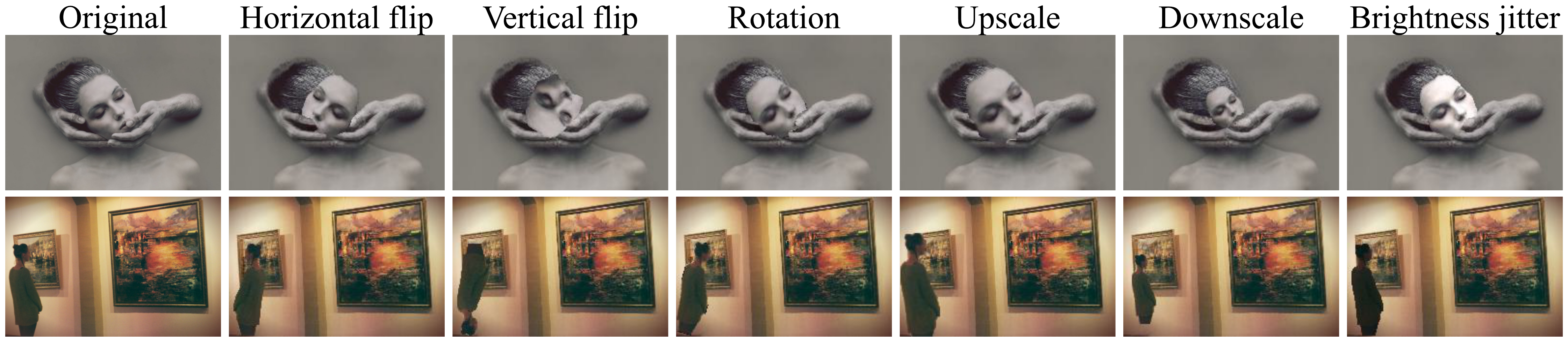}
    \caption{BackFlip with different local transformations on images from TAD66K - Art.}
    \label{fig:tad66k_trans}
\end{figure}

Understanding how different image transformations affect perceived aesthetic value for human observers and models is important. We consider six local image transformations using BackFlip, illustrated on example images from TAD66k - Art in  Figure~\ref{fig:tad66k_trans}. We compare the results to assess the effect of BackFlip with different local augmentation techniques using ResNet-18 in Table~\ref{tab:local_aug}. According to the results, all the local augmentations perform similarly. However, upscale appears to yield higher PCC and SRCC scores, likely because the augmented segments cover more of the inpainted background, thereby mitigating the loss of information due to inpainting. In terms of accuracy, downscale performs slightly better. 

\begin{table}[t]
\fontsize{7}{9}\selectfont
\centering
\begin{tabular}{l|lll}
\hline 
\textbf{Local augmentation} & \textbf{Acc. (\%)} & \textbf{PCC} & \textbf{SRCC}  \\
\midrule 
Horizontal flipping & $ 53.11 \pm 5.83 $ & $ 0.18 \pm 0.13 $ & $ 0.15 \pm 0.11 $ \\
Vertical flipping & $ 54.33 \pm 5.77 $ & $ 0.19 \pm 0.15 $ & $ 0.17 \pm 0.13 $ \\
Hor./Ver. flipping & $ 54.86 \pm 1.94 $ & $ 0.24 \pm 0.05 $ & $ 0.21 \pm 0.04 $ \\
Rotation & $ 53.34 \pm 5.06 $ & $ 0.20 \pm 0.14 $ & $ 0.18 \pm 0.12 $ \\
Upscale & $ 54.33 \pm 2.45 $ & $ \mathbf{0.25 \pm 0.02} $ & $ \mathbf{0.23 \pm 0.03} $ \\
Downscale & $ \mathbf{55.64 \pm 1.28} $ & $ 0.24 \pm 0.01 $ & $ 0.20 \pm 0.01 $ \\
Brightness jitter & $ 54.19 \pm 3.78 $ & $ 0.22 \pm 0.05 $ & $ 0.19 \pm 0.06 $ \\
\hline
\end{tabular}
\caption{Testing the effect of BackFlip with different local augmentation methods on TAD66K - Art using ResNet-18. We report the average across 5 independent runs. 
}
\label{tab:local_aug}
\end{table}

\subsubsection{Number of augmented segments.} 
\begin{figure}[!ht]
    \centering
    \includegraphics[width=0.9\textwidth]{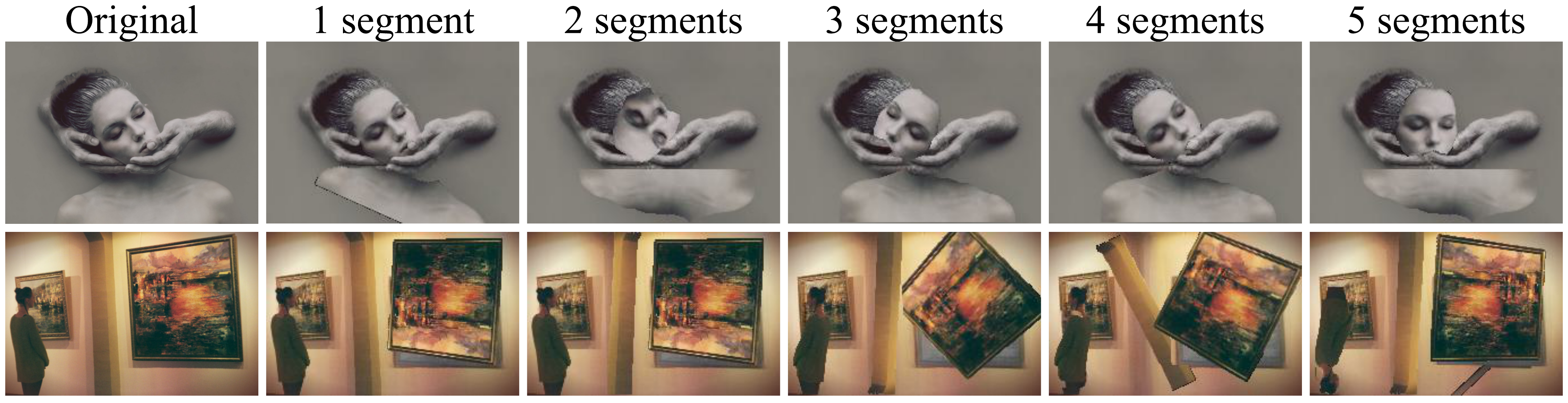}
    \caption{Local image transformations (rotation, horizontal and vertical flip) using BackFlip with increasing number of segments. Images from the TAD66k - Art dataset.}
    \label{fig:tad66k_seg}
\end{figure}

We illustrate the effect of the number of segments on images from the TAD66k - Art dataset in Figure~\ref{fig:tad66k_seg}. This figure shows the effects of three local image transformations (rotation, horizontal and vertical flip) using BackFlip, considering up to five augmented segments. As observed, the number of segments augmented through BackFlip significantly affects the visual dissimilarity between the augmented images and the original image. 

We compare the effect of the number of augmented segments on model performance in Table~\ref{tab:n_segments}, keeping every other aspect of the training set-up constant between comparisons. We train ResNet-18, pre-trained on ImageNet, on the TAD66K - Art dataset. Parameters for the BackFlip components and the chosen inpainting method, in this case, LaMa, are fixed. We consistently use either vertical flip or horizontal flip for the local augmentations applied to each segment. The results suggest that one segment is the most optimal choice for local augmentation for artworks. However, we do not observe a significant difference between the results across different numbers of segments, except possibly for four segments. Notably, these results do not show consistent improvements across all five runs. This inconsistency could have multiple explanations. 
It could be due to varying segment sizes, which change the percentage of image alterations between runs, or the interaction between chosen segments and local augmentations (horizontal vs. vertical flip), influenced by the complex nature of artistic images.


\begin{table}[!ht]
\fontsize{7}{9}\selectfont
\centering
\begin{tabular}{l|lll}
\hline 
\textbf{Number of segments} & \textbf{Acc. (\%)} & \textbf{PCC} & \textbf{SRCC}  \\
\midrule 
1 &  $\mathbf{55.19 \pm 2.44}$	& $\mathbf{0.26 \pm 0.04}$	& $\mathbf{0.23 \pm 0.04}$ \\
2 &  $54.5 \pm 3.42$	& $0.24 \pm 0.04$	& $0.21 \pm 0.04$ \\
3 & $54.86 \pm 1.94$	& $0.24 \pm 0.05$	& $0.21 \pm 0.04$ \\
4 &  $52.8 \pm 3.06$	& $0.24 \pm 0.04$	& $0.21 \pm 0.04$ \\
5 &  $54.6 \pm 2.48$	& $0.23 \pm 0.04$	& $0.21 \pm 0.05$ \\
\hline
\end{tabular}
\caption{The effect of the number of augmented segments on BackFlip. The results are obtained on the TAD66k - Art dataset using ResNet-18. We report the average across 5 independent runs. As explained in Section~\ref{segmentation_section}, we exclude the segments covering more than 90\% of the image and select the remaining segments in descending size.
}
\label{tab:n_segments}
\end{table}\label{ablation}

\section{Conclusion and Future Work}
We introduce BackFlip and examine the impact of local and global data augmentation on artistic IAA. Local augmentations, such as BackFlip, preserve the overall composition of images while introducing variations that do not affect global aesthetic qualities, making them advantageous for artistic IAA. Our experiments demonstrate that local augmentations  outperform global ones in the majority of our tests. A notable contribution of our study is the inclusion of the erase+inpainting technique within the BackFlip pipeline, as well as BoxFlip, which further enhances the effectiveness of local augmentations. 
This underscores the importance of local augmentations that preserve overall composition and the crucial role that composition plays in the aesthetics of artworks.


Additionally, we emphasize that the dataset quality plays a crucial role in artistic IAA. A well-curated and diverse dataset is essential for reliable results. We observed that the annotations in some datasets, such as BAID, are too noisy to provide a good supervisory signal. Furthermore, the images in TAD66k, which include pictures of artworks, frames, graffiti, and even nail art, do not always align with the typical global aesthetic qualities expected in paintings.

Our findings suggest that BackFlip is a promising technique for artistic IAA and holds potential for broader applications in aesthetics research. An interesting future work would be to explore the impact of varying parameter settings across different local augmentation methods. Deploying BackFlip can facilitate empirical aesthetics research by collecting ratings from human participants for augmented data, further enriching our understanding of aesthetic evaluation.


\noindent
\textbf{Acknowledgments}
Funded by the European Union (ERC AdG, GRAPPA, 101053925, awarded to Johan Wagemans). Views and opinions expressed are however those of the authors only and do not necessarily reflect those of the European Union or the European Research Council Executive Agency. 
We acknowledge the insightful discussions with Lisa Koßmann and Hayley Hung. 

%
%
\bibliographystyle{splncs04}
\bibliography{main}
\end{document}